\pdfoutput=1
\documentclass{article}

     \usepackage[final, nonatbib]{neurips_2019}

\usepackage[utf8]{inputenc} 
\usepackage[T1]{fontenc}    
\usepackage[numbers]{natbib}
\usepackage{hyperref}

\usepackage{url}            
\usepackage{booktabs}       
\usepackage{amsfonts}       
\usepackage{nicefrac}       
\usepackage{microtype}      

\usepackage{amsmath}
\usepackage{graphicx}
\usepackage{gensymb}

\usepackage{subfigure}
\usepackage{graphicx}
\usepackage{pdfpages}
\usepackage{lipsum}
\usepackage{float}

\newcommand\blfootnote[1]{%
  \begingroup
  \renewcommand\thefootnote{}\footnote{#1}%
  \addtocounter{footnote}{-1}%
  \endgroup
}

\title{Deep Learning in Spiking Phasor Neural Networks}

\author{%
  Connor Bybee \\
  Computational Biology \& Redwood Center\\
  UC Berkeley\\
  Berkeley, CA 94720 \\
  \texttt{bybee@berkeley.edu} \\
   \And
   E. Paxon Frady \\
   Redwood Center \& Intel \\
   UC Berkeley\\
   Berkeley, CA 94720 \\
   \texttt{epaxon@berkeley.edu} \\
   \AND
   Friedrich T. Sommer \\
   Redwood Center \& Intel\\
   UC Berkeley \\
   Berkeley, CA 94720 \\
   \texttt{fsommer@berkeley.edu} \\
}

\begin{document}

\maketitle

\begin{abstract}
  Spiking Neural Networks (SNNs) have attracted the attention of the deep learning community for use in low-latency, low-power neuromorphic hardware, as well as models for understanding neuroscience. In this paper, we introduce Spiking Phasor Neural Networks (SPNNs). SPNNs are based on complex-valued Deep Neural Networks (DNNs), representing phases by spike times. Our model computes robustly employing a spike timing code and gradients can be formed using the complex domain. We train SPNNs on CIFAR-10, and demonstrate that the performance exceeds that of other timing coded SNNs, approaching results with comparable real-valued DNNs.
\end{abstract}
\blfootnote{Work presented at Intel’s Neuromorphic Community Fall 2019 workshop in Graz, Austria and the UC Berkeley Center for Computational Biology Retreat 2019 }

\section{Introduction}

The highest classification accuracy on image recognition tasks has been achieved by real-valued deep neural networks (DNN) with floating point precision. But recent effort has attempted to reformulate deep learning into formats more suitable for hardware acceleration and low-energy edge computing. This work aims to demonstrate the benefits of using spike timing codes in a way that can advantage deep learning acceleration according to specific metrics, especially energy efficiency and latency.

Implementation of DNN algorithms on next-generation computing platforms such as neuromorphic computer architectures and networks of coupled-oscillators is promising, and may prove to be more energy-efficient and faster than current hardware based on the Von Neumann architecture \cite{schuman2017survey, nikonov2015coupled, davies2018loihi}. When mapping DNN neurons to spiking neural networks (SNN), the machine learning and neuroscience community have largely focused on the use of rate codes \cite{hunsberger2015spiking, lee2016training, roy2018learning, o2018training}. Spiking networks are typically challenged by a non-differentiable objective function, which prohibits training with backpropagation. Successful methods include training ANNs with the purpose of converting them to SNNs, using spiking versions of backpropagation compatible with spike timing-dependent-plasticity, and use of other learning rules \cite{lee2016training, mostafa2018supervised}. This paper presents a method for using spike events to represent complex-valued neural states, which can easily be differentiated in the complex domain.

In a rate code, a stimulus feature is represented by the firing rate of a neuron measured by the count of spikes in an extended time window. Conversely, in a spike timing code a feature is represented by the precise timing of a single spike. Potential benefits of spike timing codes include (i) efficiency in terms of the number of spikes per computation, and (ii) computation speed, since the duration of a step of neural update with a rate code cannot be shorter than the integration window required for estimating the lowest rate. On the other hand, spike timing codes are often regarded as brittle and not suited for implementing robust computations \cite{london2010sensitivity}. 

Recently, Thresholding Phasor Associative Memories (TPAM) were introduced \cite{frady2019robust} that illustrate how a complex-valued neural state can be directly mapped to a spike timing code. Essentially, relative to an ongoing oscillation, the timing of the spike can be used to indicate the phase of a complex number. This was shown to provide robust spike timing codes in the context of attractor neural networks. Here, we extend this timing code to feed forward deep networks, and show how backpropagation based learning can be applied in the complex domain and robustly mapped to SNNs.  Motivated by that aim, this paper compares and demonstrates deep learning in spiking phasor networks on benchmarks MNIST \cite{lecun1998gradient} and  CIFAR-10 \cite{krizhevsky2009learning}.




\section{Background}

\subsection{Spiking backpropagation}

Supervised learning of spike timing codes was explored in \cite{mostafa2018supervised}. In \cite{mostafa2018supervised}, neurons are trained to integrate input and fire at a precise time upon crossing a threshold. Training is performed by first determining the set of input spikes leading to an output spike, called the causal set. Parameters updates are then calculated by backpropagating through the causal set. Our work presents a method to train a spike timing coded SNN without the need to calculate a casual set. 

A method for converting real-valued ANNs to timing coded SNNs was introduced in \cite{rueckauer2018conversion}. It demonstrates on a simple machine learning task (MNIST) that spike timing codes are more efficient than rate codes and real-valued DNNs when considering the number of operations needed to perform a computation. Our paper extends spike timing coded SNNs to CIFAR-10.

\subsection{Neuroscience}

In Theoretical Neuroscience, it is an open question how a network of spiking biological neurons can precisely coordinate spikes over time and space. Izhikevich proposed a spiking network that creates ``polychronous'' spiking patterns, as a model for describing neural dynamics in brain circuits \cite{izhikevich2006polychronization}. Polychronization means that neurons exhibit time-locked spiking patterns, but with arbitrary spike timing and not just perfectly synchronous spiking. The spiking patterns we propose are also polychronous patterns, but these are generated by different mechanisms.  

Precise spike timing patterns are observed in multiple brain regions and cell types. The electrical properties of the neuron membrane give rise to sustained intrinsic spike oscillations, subthreshold oscillations, and membrane resonance \cite{izhikevich2007dynamical}. Neuron morphology and localized genetic expression patterns of membrane ion-channels can create compartments with specific electrical properties \cite{beaulieu2018enhanced, matsumoto2016ionic}. Rate and timing codes have been explored as an method of mapping ANNs to models of biological plausible neural networks \cite{scellier2017equilibrium, sacramento2018dendritic, mostafa2018supervised, o2018training}. Inspired by these observations, this work uses such oscillatory mechanisms to construct and simulate a SNNs with spike timing codes.

\section{Methods}
\subsection{Spiking Phasor Neural Networks}
Complex-valued neural networks have been explored in several contexts. Noest \cite{noest1988phasor} introduced phasor neural networks (PNN) with applications as an associative memory and suggested use with backpropagation. Complex domain backpropagation was introduced in \cite{georgiou1992complex} and \cite{hirose1992continuous}. Recent developments have been made with deep neural networks \cite{trabelsi2017deep}. We provide a concrete example of a feed forward PNN trained with backpropagation to implement spike timing codes. This paper refers to PNNs with the TPAM activation function \cite{frady2019robust} as Spiking PNNs (SPNNs). 

\begin{figure}
    \centering
    \includegraphics[width=0.5\textwidth]{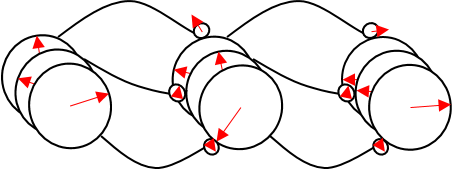}
    \caption{An illustration of a feed forward PNN with normalizing activation function or Spiking PNN (SPNN).}
    \label{fig:pnn}
\end{figure}

In PNNs, neurons are represented by complex-valued phasors $s_i \in \mathbb{C}$. Where $s_{i} = r_{i} e^{\mathrm{i} \theta_{i}} = r_{i} \cos{\theta_{i}} + \mathrm{i} r_{i} \sin{\theta_{i}} = u_i + \mathrm{i}v_{i} $. Here, $\mathrm{i} = \sqrt{-1}$. Phasors are restricted to the unit circle with the use of a non-linear activation function that preserves phase information. $\|s_i\|_2 = 1, \forall i \in \{1,..,N_s\}$

The network units typically consist of a set of input, output, and hidden variables, $\mathbf{s} = \{\mathbf{x}, \mathbf{y}, \mathbf{h}\} \in \mathbb{C}^{N_s}$. The synaptic weights and neuron biases, $W_{ij} \in \mathbb{C}$ and $b_i \in \mathbb{C}$, act as attenuating or amplifying phase shifters.  

We present a two-layer network with layer units $\mathbf{h}^{(l)} \in \mathbb{C}^{N_h}, l \in \{1, 2\}$. Input units can be real-valued but are transformed to unit length complex-valued vectors whose phase is a function of the real-valued input. $x_i = e^{\mathrm{i} \theta_i}$, where $\theta_i = g_x(\Tilde{x}_i)$, $\Tilde{x}_i \in \mathbb{R}$. Where $g_x: \mathbb{R} \rightarrow [0, \pi]$. The output of the network is mapped to phases similar to binary phase shift keying. $ y_i = e^{\mathrm{i} \theta_i}$, where $\theta_i = g_y(\Tilde{y}_i)$, $\Tilde{y}_i \in \{0,1\}$ and $g_y: \{0,1\} \rightarrow \{0, \pi \}$.

Inference proceeds by propagating activity from the input layer to the output layer: 
\begin{equation}
    \mathbf{h}^{(l)} = f(\mathbf{W}^{(l)} \mathbf{h}^{(l-1)} + \mathbf{b}^{(l)}, \Theta)
\label{update}
\end{equation}
The vector $\mathbf{h}^{(l)}$ is the activity of hidden layer $l$, and for computing the activity in the first hidden layer, equation (\ref{update}) is applied to the input vector, i.e., $\mathbf{h}^{(0)} = \mathbf{x}$. The element-wise thresholding and normalizing function is given by: 
\begin{align}
    f(z_i, \Theta) = 
    \begin{cases}
    \displaystyle
    \frac{z_i}{|z_i|}  & \text{if } |z_i| - \Theta > 0 \\
    0 & \text{otherwise}
    \end{cases}
    \label{nonlinearity}
\end{align}
where $\Theta$ is a trainable thresholding parameter.

The objective is to minimize the phase difference between the network output and targets. In supervised learning tasks, often the softmax with cross-entropy loss is used with the target variables are encoded as one-hot binary vectors. Here, we encode class labels onto the unit circle. Positive class labels are encoded to lead negative class labels in phase. A  classification is successful when the phase of the positive class leads the phase of all other outputs. The class targets are binary variables encoded onto the unit circle, similar to binary phase shift keying. Positive and negative classes are out of phase by $180 \degree$ or $\pi$ radians. The loss function is given by:
\begin{equation}\label{objective}
    L = \frac{1}{2} \|\mathbf{y} - \hat{\mathbf{y}}\|^2_2 = N_y - \sum_{i=1}^{N_y} \cos(\theta_i - \hat{\theta}_i) 
\end{equation}
Where $\theta_i$ are the target phase angles and $\hat{\theta}_i$ are the estimated phase angles. Minimizing the objective corresponds to phase aligning the output and targets:
\begin{equation}\label{grad}
    \frac{ \partial L }{ \partial \hat{\theta}_i} = \sin(\theta_i - \hat{\theta}_i)
\end{equation}
As we will show in the results, PNNs with the TPAM activation function can be trained with the backpropagation algorithm to minimize equation \ref{objective}. 

\subsection{Complex domain backpropagation} 
In \cite{georgiou1992complex}, complex domain backpropagation was derived for the activation function below. 

\begin{align*}
     y &= f(z) \\
    &= \frac{z}{c + \frac{1}{r}|z|}\\
    &= u + \mathrm{i}v 
\end{align*}

Where, $z = a + \mathrm{i}b \in \mathbb{C}$, $\{a, b\} \in \mathbb{R} $ and  $|z| = \sqrt{a^2 + b^2}$. The partial derivatives are
\begin{align*}
&\frac{\partial u}{\partial a} = 
    \begin{cases}
    \frac{r(b^2 + cr|z|)}{|z|(cr + |z|)^2} & \text{if } |z| > 0 \\
    \frac{1}{c} & \text{if } |z| = 0
    \end{cases}\; \; \; \; \;
    \; \frac{\partial u}{\partial b} = 
    \begin{cases}
    -\frac{rab}{|z|(cr + |z|)^2} & \text{if } |z| > 0 \\
    0 & \text{if } |z| = 0
    \end{cases}\\
    &\frac{\partial v}{\partial a} = 
    \begin{cases}
    -\frac{rab}{|z|(cr + |z|)^2} & \text{if } |z| > 0 \\
    0 & \text{if } |z| = 0
    \end{cases} \; \; \; \; 
    \frac{\partial v}{\partial b} = 
    \begin{cases}
    \frac{r(a^2 + cr|z|)}{|z|(cr + |z|)^2} & \text{if } |z| > 0 \\
    \frac{1}{c} & \text{if } |z| = 0
    \end{cases}
\end{align*}

Considering the case where $\Theta=0$ and $|z| > 0$, the SPNN activation function is found by setting $r=1$ and in $\lim_{c \to 0}$. The partial derivative can be written in the complex domain by adding together the individual terms.

\begin{align*}
 (\frac{\partial u}{\partial a} + \frac{\partial v}{\partial a}) + \mathrm{i} (\frac{\partial u}{\partial b} + \frac{\partial v}{\partial b}) = \frac{(b^2 - a b) + i (a^2 - a b)}{(a^2 + b^2)^{3/2}}
\end{align*}

\subsection{Spiking neuron model}
The continuous-valued phasor network can be mapped to discrete spiking events in time. For a given choice $T$ for the length of a cycle, the each phase angle, $\theta_i \in [0, 2\pi]$ can be mapped to a time $t_i = \theta_i * T/2\pi$. 

We first consider a network where all layers are phase-locked to the input layer. The real-valued inputs are mapped to unique phases on the unit circle. This creates an ambiguity between the largest and smallest values since $2 \pi$ is equivalent to $0$. Therefore, we limited the values to $[0, \pi]$. We chose this option since it is simple and is sufficient for demonstration purposes. Each input phasor can also be assigned a fixed, random phase shift. 

In the first cycle, the input units spike and each synapse is activated with magnitude $|W_{ij}|$ after a delay of $\theta_{ij}*T/2\pi$. Each neuron integrates the information from the current cycle and spikes after the appropriate delay during the subsequent cycle. Therefore, each cycle propagates signals to adjacent layers, see figure \ref{fig:phase_codes}.

\subsection{Circuit Model}
We simulate a circuit model to demonstrate that SPNNs can be implemented by simple electrical components and are robust to noise induced by numerical simulation and changing input. A change in soma membrane potential is driven by a leakage current and combined input from dendrites: 
\begin{align*}
    {\frac {{\mathrm {d} }V_{m}}{{\mathrm {d} }t}} &= \frac{g_{l}(V_{l}-V_{m}) + g_{c}(V_{d} - V_{m} - \overline{V}_{d})} {C_{m}} \\
    {\frac {{\mathrm {d} }\overline{V}_{d}}{{\mathrm {d} }t}} &= \frac{(V_{d} - \overline{V}_{d})} {\tau_{d}}
\end{align*}

$V_{m}$ is the membrane potential, $g_{l}$ is the conductance of the leak channel, $V_{l}$ is the leak reversal potential, $g_{c}$ is the conductance of between the soma and dendrite, $V_{d}$ is the dendrite potential, $\overline{V}_{d}$ is the average dendrite potential, $\tau_{d}$ is average dendrite potential time constant, and $C_m$ is the membrane capacitance. To correct for bias around the threshold voltage, the average dendrite potential is subtracted from the instantaneous dendrite membrane potential. Dendrites integrate current from synapses which generate membrane potential oscillations after a spike arrives:
\begin{align*}
    V_{d} &= \sum_{i}w_{i} V_{s, i}\\
    \frac{\mathrm {d} V_{s, i}}{\mathrm {d}t} &= \frac{-W_{s, i}} {C_{m}}\\
    \frac{\mathrm {d} W_{s, i}}{\mathrm {d}t} &= \frac{V_{s, i}}{L} - \frac{W_{s, i}} {\tau_{s}}
\end{align*}
$V_{s, i}$ is the potential at synapse $i$, $w_i$ is the synaptic weight. $W_{s, i}$ are activation parameters which act to generate membrane potential oscillations at a specific frequency, $L$ is a constant used to tune the membrane potential frequency, and $\tau_{s}$ is the synaptic time constant which controls oscillation damping. A spike resets the synapse parameters such that $V_{s, i} = 0$ and $W_{s, i} = W_{spike}$. The neuron spikes when the soma membrane potential cross a threshold. The refractory period is maintained until the neuron soma membrane potential becomes negative.

\section{Results}
\begin{figure}[h]
    \centering
    \subfigure[]{\label{fig:pc_a}\includegraphics[width=.49\textwidth]{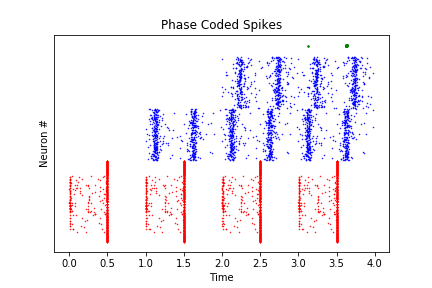}}
    \subfigure[]{\label{fig:pc_b}\includegraphics[width=.49\textwidth]{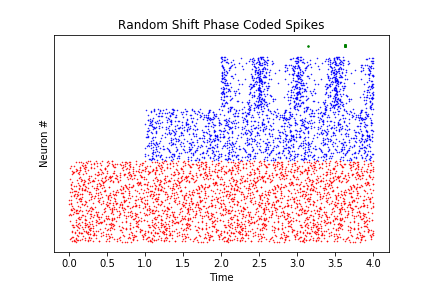}}
    \caption{The first three cycles of inference for a feed forward SPNN trained on MNIST. The network is phased locked to the input signal. Each cycle allows information to propagate to the next layer. Each line in the vertical axis is a neuron. The input, hidden, and output layers are colored red, blue, and green, respectively. The output neurons are accurately separated in phase. The input pixel amplitudes are mapped to phase angels inversely proportional to the magnitude. \ref{fig:pc_a}: Similar input amplitudes are phase locked. \ref{fig:pc_b}: For demonstration, each input can have a fixed phase shift while still maintaining performance. }
    \label{fig:phase_codes}
\end{figure}
\begin{figure*}[!h]
    \subfigure[]{\label{fig:a}\includegraphics[width=0.5\textwidth]{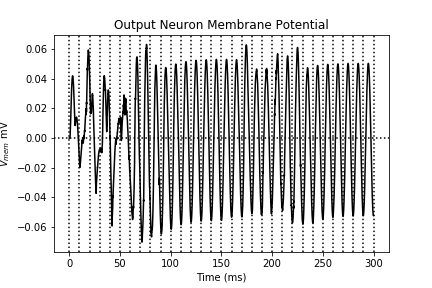}}
    \subfigure[]{\label{fig:b}\includegraphics[width=0.5\textwidth]{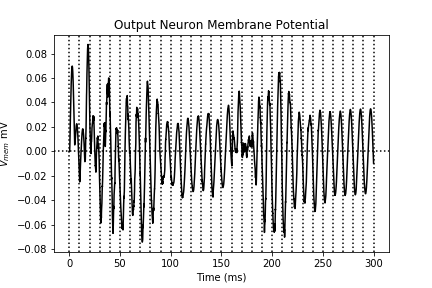}}
    \subfigure[]{\label{fig:c}\includegraphics[width=0.5\textwidth]{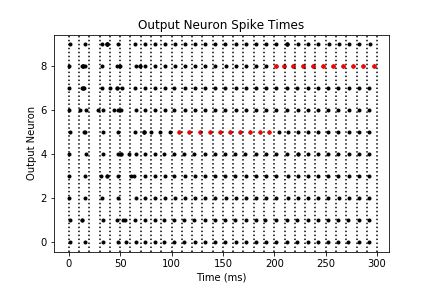}}
    \subfigure[]{\label{fig:d}\includegraphics[width=0.5\textwidth]{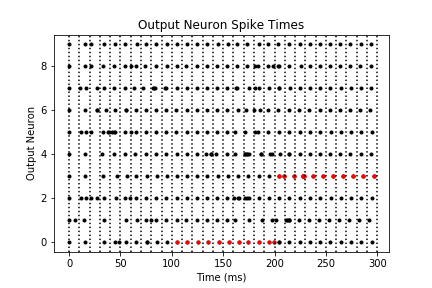}}
    \subfigure[]{\label{fig:e}\includegraphics[width=0.5\textwidth]{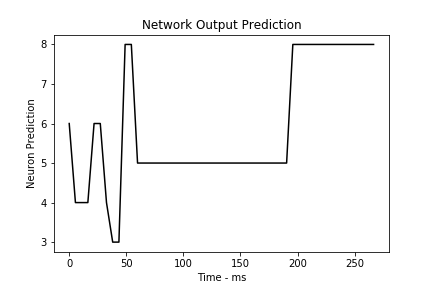}}
    \subfigure[]{\label{fig:f}\includegraphics[width=0.5\textwidth]{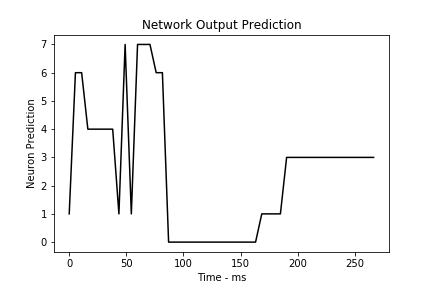}}

    \caption{Simulation of the circuit model SPNN on MNIST (left) and CIFAR (right). Left: An example with label 5 is presented for 15 cycles or 150ms. Immediately following, an example with label 8 is presented for 15 cycles. \ref{fig:a}: The membrane potential for a unit in the output layer. The unit is unstable until cycle 7 when it beings to lock to the input signal. \ref{fig:b}: Spike times for the output layer. \ref{fig:e}: Predicted class label over time. Right: An example with label 0 is presented for 15 cycles or 150ms. Immediately following, an example with label 3 is presented for 15 cycles. \ref{fig:b}: The membrane potential for a unit in the output layer. The unit is unstable until cycle 9 when it beings to lock to the input signal. \ref{fig:d}: Spike times for the output layer. \ref{fig:f}: Predicted class label over time.}
    \label{fig:snn}
\end{figure*}

For visualization purposes, we train a fully connected SPNN on MNIST, 784-512-512-10. We unroll the network activity in time to demonstrate how signals propagate (Fig.~\ref{fig:phase_codes} - left panel). The spike times represent a stable limit-cycle. For demonstration purposes, we add a fixed, random phase shift to each component of the input while maintaining correct classification (Fig.~\ref{fig:phase_codes} - right panel). A system that has yet to reach the limit-cycle will have a settling time (transient settling dynamics visible later in Fig.~ \ref{fig:snn}).

Convolutional SPNNs are trained on MNIST and CIFAR-10. The convolutional networks consist of two layers of convolution without padding followed by three fully connected layers,  conv(6,3x3)$\to$conv(16,3x3)$\to$FC128$\to$FC128$\to$FC10. No normalization or pooling is used. We train using Pytorch \cite{paszke2017automatic}. The Adam optimizer \cite{kingma2014adam} is used with default parameters and a learning rate of 0.001. We train small networks compared to state-of-the-ar                                                                                                                                                                                                                                         t for simulation purposes. Future work will investigate the performance of SPNNs on more difficult tasks and with larger networks. 
 
\begin{figure}[h]
    \centering
    \subfigure[]{\label{fig:acc_a}\includegraphics[width=.49\textwidth]{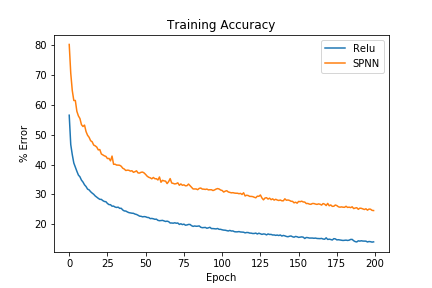}}
    \subfigure[]{\label{fig:acc_b}\includegraphics[width=.49\textwidth]{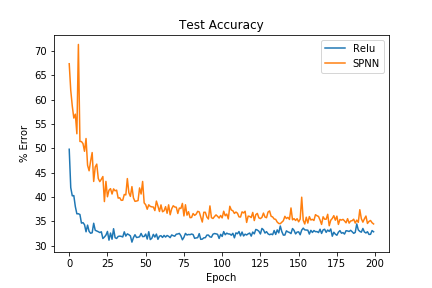}}
    \caption{Training and test accuracy on CIFAR-10. The minimum test error rates for the relu network and the SPNN network are $30.73\%$ and $34.09\%$, respectively.}
    \label{fig:cifar_acc}
\end{figure}

Training and test error rates with the convolutional SPNNs were measured for classifying MNIST data (Fig.~\ref{fig:cifar_acc}). Our results were obtained with 32-bit floating point precision, and therefore should be regarded as an upper-bound of the performance SPNN hardware implementations are expected to achieve using parameters from a trained model. The classification precision our model achieved surpasses the best results previously reported for SNNs with spike timing codes \cite{rueckauer2018conversion}. Further, SPNNs approach the performance level of comparable DNNs, our model comes within $0.02\%$ compared to the error rate of the DNN used for conversion to a SNN. Last, our network outperforms previous methods for directly training SNNs with spike timing codes \cite{mostafa2018supervised}. 
\begin{figure}[h]
    \centering
    \includegraphics[width=.49\textwidth]{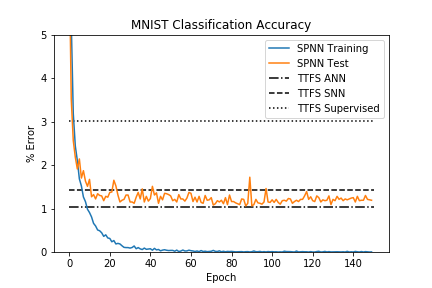}
    \caption{Results for a SPNN on MNIST. The lowest test error rate previously reported for SNNs with spike timing codes and the ANN used for conversion to the SNN is $1.43\%$ and $1.04\%$, respectively \cite{rueckauer2018conversion}. The lowest test error rate previously reported for supervised training of spike timing codes is $3.03\%$ \cite{mostafa2018supervised}. SPNNs achieve a test error rate of $1.06\%$.  }
    \label{fig:mnist_acc}
\end{figure}

When trained on CIFAR-10. To our knowledge, there is no previous reports of performing supervised learning and inference with spike timing codes on CIFAR-10. However, due to the rather small network size we used, our results on training and test error rates do not reach the state-of-the-art (Fig.~\ref{fig:cifar_acc}). It is promising that SPNN approach the performance of real-valued DNNs on CIFAR despite minimizing the mean-squared error opposed to the softmax with cross-entropy loss. 


We performed simulations of convolutional SPNNs implemented with Brian2 \cite{goodman2009brian}, an open-source simulator for spiking neural networks. The simulations of SPNNs in the circuit are modeled at time scales and firing rates relevant to spiking biological neurons, but the simulation time scale is arbitrary. For the parameter values, $g_{l} = \pi C_m /T$, $g_{c} = 60\pi C_m /T$,  $V_{l}=0$mV, $C_m=10$pF, $L=\frac{1}{(2 \pi / T)^2 C_m}$, $W_{\text{spike}}=.3$pA, $\tau_{d}=0.8T$, and $\tau_s =0.0T$. Simulations are for 300ms. The cycle period, $T$, is 10ms with a frequency of 100Hz. Numerical integration is performed with forward euler. The time step is 0.025ms. The period to time step ratio is approximately 400 to 1. The first example is presented for 15 cycles or 150ms (Fig.~\ref{fig:snn}). The second example is then presented for 15 cycles. The output prediction is calculated as the unit which is furthest out of phase with respect to the other output units within a 3 cycle window, $ \text{argmax}_{i} \ \mathbb{E}_{t_{spike, i}} \frac{1}{2}(|t_{\text{next spike}, j \neq i} - t_{spike, i}| + |t_{\text{last spike}, j \neq i} - t_{spike, i}|)$. Here, the expectation is taken with respect to the spike times of each component $i$, within a three cycle window.

At this resolution, we notice a decrease in performance compared to the original SPNN. Currently, simulating all MNIST and CIFAR test samples is prohibitively expensive. We report qualitative results, leaving a quantitative analysis of the SPNN performance versus numerical precision for future work. The performance gap decreases as the time step decreases relative to the cycle period and as the confidence in the output prediction increases. It is promising that even at this resolution SPNN are robust to noise in the spike times.

The phenomenon of pipelining emerges in SPNNs as an effect of a changing input stimulus (Fig.~\ref{fig:snn}). Pipelining is a technique used for reducing the latency and increasing the resource utilization of computing hardware by connecting processing elements in series and passing the output of one unit to the input of the next unit in the pipeline. The signals propagate similar to travelling waves in one direction.

\section{Discussion}
With the advent of powerful hardware solutions (TrueNorth \cite{akopyan2015truenorth}, Loihi \cite{davies2018loihi}) for spiking neurons, the question arises how deep learning can be implemented with spikes. The current approach is to implement real-valued activation values by spike rate \cite{hunsberger2015spiking, lee2016training}. Some of these approaches yield near state-of-the-art results but require a large number of spikes and exhibit rather long response latency (time to solution) due to the fact that the estimation of small rates requires large integration windows. A potential remedy to this problem is to compute with spike times \cite{rueckauer2018conversion}, but current solutions are brittle and cannot be scaled up to challenging machine learning problems. 

Here, we present SPNNs employing spike timing codes and show them to be successful at solving challenging machine learning tasks. In our SPNN model, spike times
represent phase angles of a complex number. Upon receiving input, the network computes robustly by relaxing to a stable limit-cycle of periodic spike trains, which is a fixed-point in the complex domain.
Our work is probably the first demonstration how a SNN employing a spike timing code can learn the CIFAR-10 dataset. 

Another problem that our model can address is how to implement gradient learning in spiking neural networks using local signals. If deep learning is implemented by a rate code, pre- and post-synaptic spikes need to be integrated to estimate the activity values required in the gradient-based learning rule. In contrast, in the SPNN the timing of spikes exactly encode the continuous-valued phase variables, which are required in the learning rule.  

Our model is based on a phasor network with a complex-valued normalizing activation function, first proposed in  \cite{frady2019robust}, constraining the phasor variables to binary magnitude values of 0 or 1. It may be possible to extend our model to combine rate and timing codes by representing continuous-valued magnitudes of phasor variables in spike rates. In addition, there is other recent work on deep learning in complex networks \cite{trabelsi2017deep}, proposing activation functions different from the one used here. Future work should explore these other activation functions in the context of SPNNs.

Our work has also interesting implications for neuroscience. As demonstrated in the results, SPNNs exhibit time-locked, but not necessarily synchronous spiking. This phenomenon is similar to polychronous spike patterns \cite{izhikevich2006polychronization}. Therefore, SPNNs may serve as a model for how to compute with such spike patterns, which also have been observed in neural activity \cite{tingley2018transformation}. 

\clearpage
\bibliographystyle{abbrv}

\bibliography{citations}

\end{document}